%% file: example_paper.tex
\documentclass{article}
\usepackage{microtype}
\usepackage{graphicx}
\usepackage{subcaption}
\usepackage{booktabs}
\usepackage{hyperref}

\usepackage[preprint]{icml2026}

\usepackage{xspace}
\def\ourmodel{\textbf{HIVE}\xspace}
\usepackage{amsmath}
\usepackage{amssymb}
\usepackage{mathtools}
\usepackage{amsthm}
\usepackage[dvipsnames]{xcolor}
\usepackage{booktabs}
\usepackage{xcolor}
\usepackage{pifont}
\usepackage[table]{xcolor}
\usepackage{makecell}
\usepackage[utf8]{inputenc}
\usepackage{listings} 
\usepackage{diagbox}
\usepackage{multirow}
\usepackage{tabularray}
\UseTblrLibrary{diagbox}

\definecolor{P}{HTML}{FFB2B2}
\definecolor{R}{HTML}{FFE2C8}
\definecolor{C}{HTML}{FFC492}
\newcommand{\cmark}{\textcolor{green!60!black}{\ding{51}}}
\newcommand{\xmark}{\textcolor{red!70!black}{\ding{55}}}

\usepackage[capitalize,noabbrev]{cleveref}

\theoremstyle{plain}

\theoremstyle{definition}

\theoremstyle{remark}

\usepackage[textsize=tiny]{todonotes}

\usepackage[english]{babel}
\usepackage[autostyle, english = american]{csquotes}
\MakeOuterQuote{"}
\icmltitlerunning{Multimodal Latent Reasoning via Hierarchical Visual Cues Injection}

\begin{document}

\twocolumn[
  \icmltitle{Multimodal Latent Reasoning via Hierarchical Visual Cues Injection}
  \icmlsetsymbol{equal}{*}

\begin{icmlauthorlist}
\icmlauthor{Yiming Zhang}{ntu}
\icmlauthor{Qiangyu Yan}{noah}
\icmlauthor{Borui Jiang}{noah}
\icmlauthor{Kai Han}{noah}

\end{icmlauthorlist}
\icmlaffiliation{ntu}{Nanyang Technological University}
\icmlaffiliation{noah}{Huawei Noah's Ark Lab}

\icmlcorrespondingauthor{Kai Han}{kai.han@huawei.com}

\icmlkeywords{Machine Learning, ICML}

\vskip 0.3in
]

\printAffiliationsAndNotice{}

\input{sec/0_abstract}    
\input{sec/1_intro}
\input{sec/2_related_work}
\input{sec/3_architecture}
\input{sec/4_experiments}
\input{sec/5_conclusion}

\section*{Impact Statement}

This paper presents \ourmodel, a framework designed to advance the field of MLLMs through recursive latent-space reasoning and hierarchical visual integration. The potential broader impacts of this work are summarized as follows:

Computational Efficiency and Sustainability: By performing reasoning within the latent space and utilizing a looped transformer architecture, \ourmodel reduces the reliance on extremely long text sequences (CoT) and massive parameter scaling. This contributes to more computationally efficient AI systems, potentially lowering the energy consumption and carbon footprint associated with deploying high-performance reasoning models.

Enhanced Decision Support: The integration of hierarchical visual information allows for more robust interpretation of complex scenes. This could have positive societal applications in fields requiring nuanced visual-logical analysis, such as assistive technologies for the visually impaired, medical imaging interpretation support, and autonomous system safety.

Ethical Considerations: As with all large-scale multimodal models, there is a risk that the model may inherit or amplify biases present in the training data (e.g., InternViT or large-scale text corpora). Furthermore, enhanced reasoning capabilities could be misused for generating sophisticated misinformation. We encourage the community to apply standard rigorous bias-detection and safety-filtering protocols when deploying recursive latent reasoning frameworks.

Overall, our work aims to make complex multimodal reasoning more efficient and structurally grounded, and we do not foresee any specific negative societal consequences that uniquely distinguish our research from general advancements in the field of Machine Learning.

\bibliography{example_paper}
\bibliographystyle{icml2026}

\newpage
\appendix
\onecolumn
\section{Training Dataset}
\begin{figure*}[h]
  \centering
  \includegraphics[width=\linewidth]{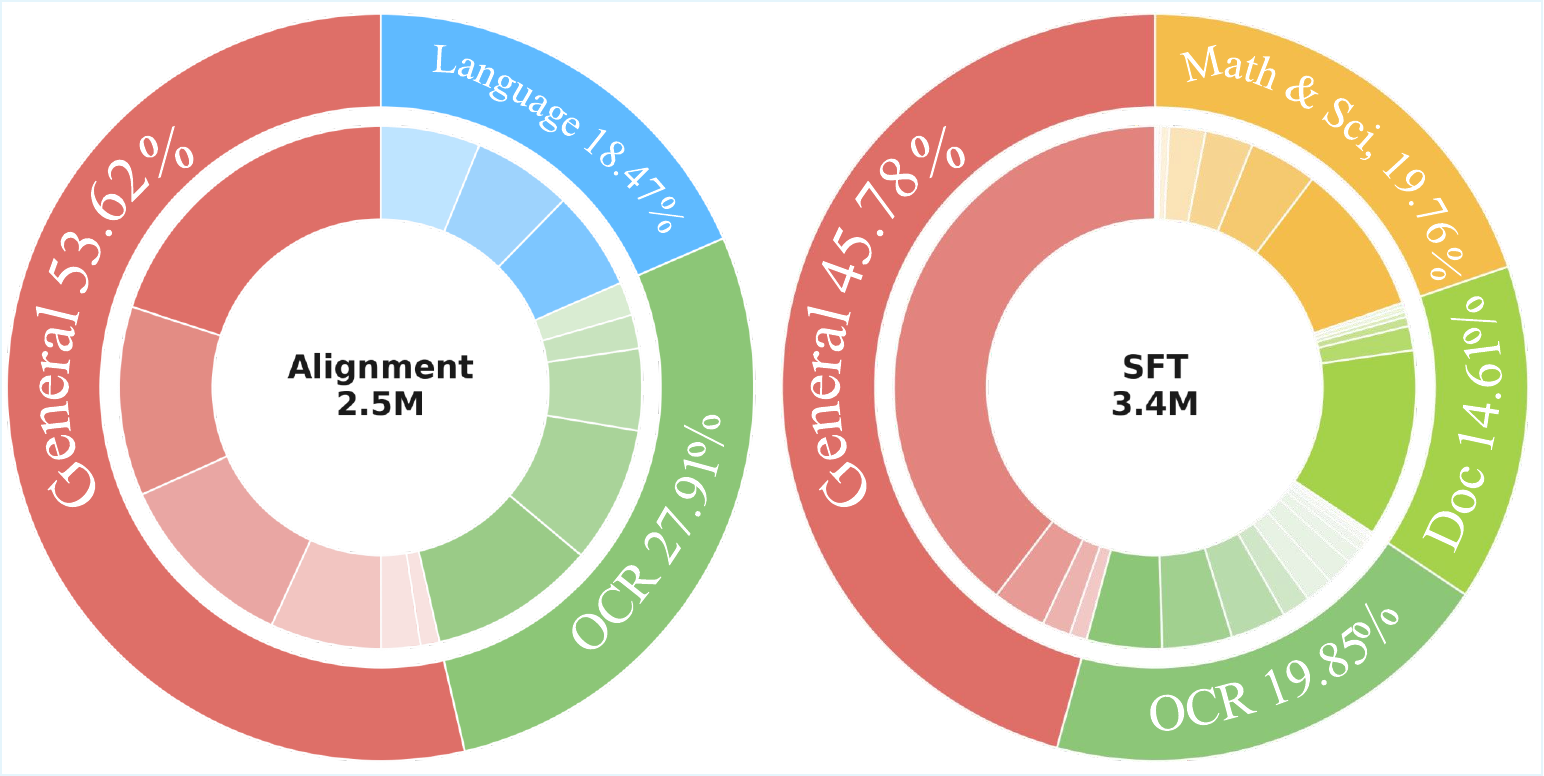}
  \caption{The details of our training datasets.}
  \label{fig:data_stat}
\end{figure*}

We provide a comprehensive breakdown of the data sources and distributions used during the training process. Detailed statistics are illustrated in Figure \ref{fig:data_stat}.

\section{Latent Space Visualizations}
\begin{figure*}[h]
  \centering
  \includegraphics[width=\linewidth]{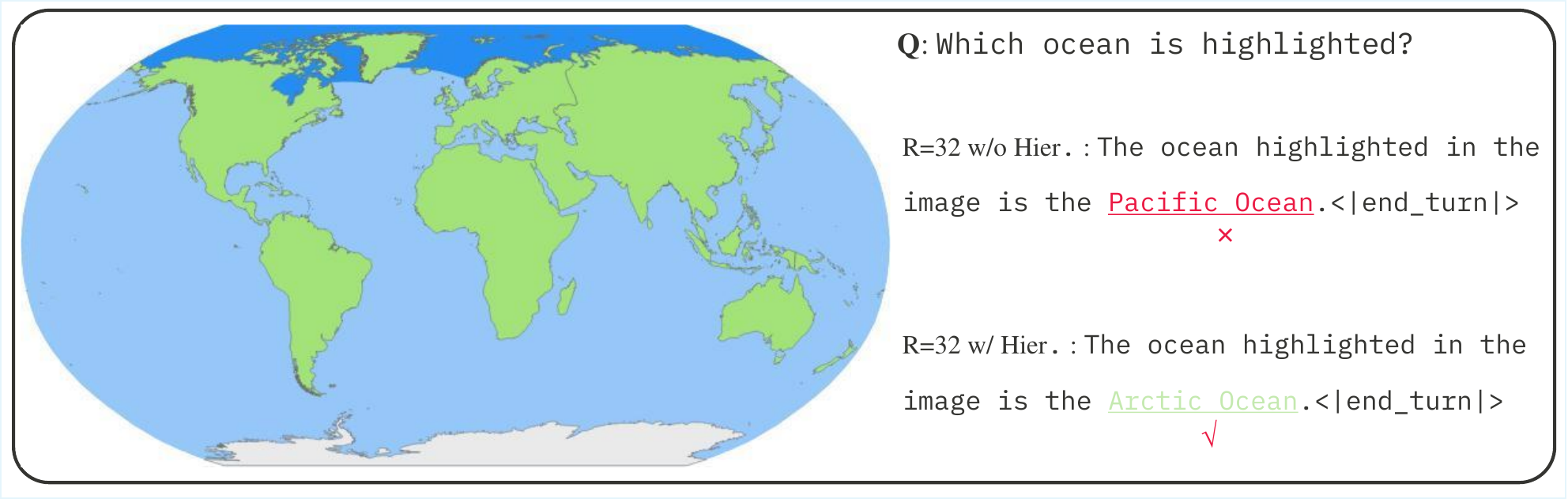}
  \caption{A case from ScienceQA, where the question has been modified as a QA.}
  \label{fig:case}
\end{figure*}

\begin{figure}
  \centering
  \includegraphics[width=0.8\linewidth]{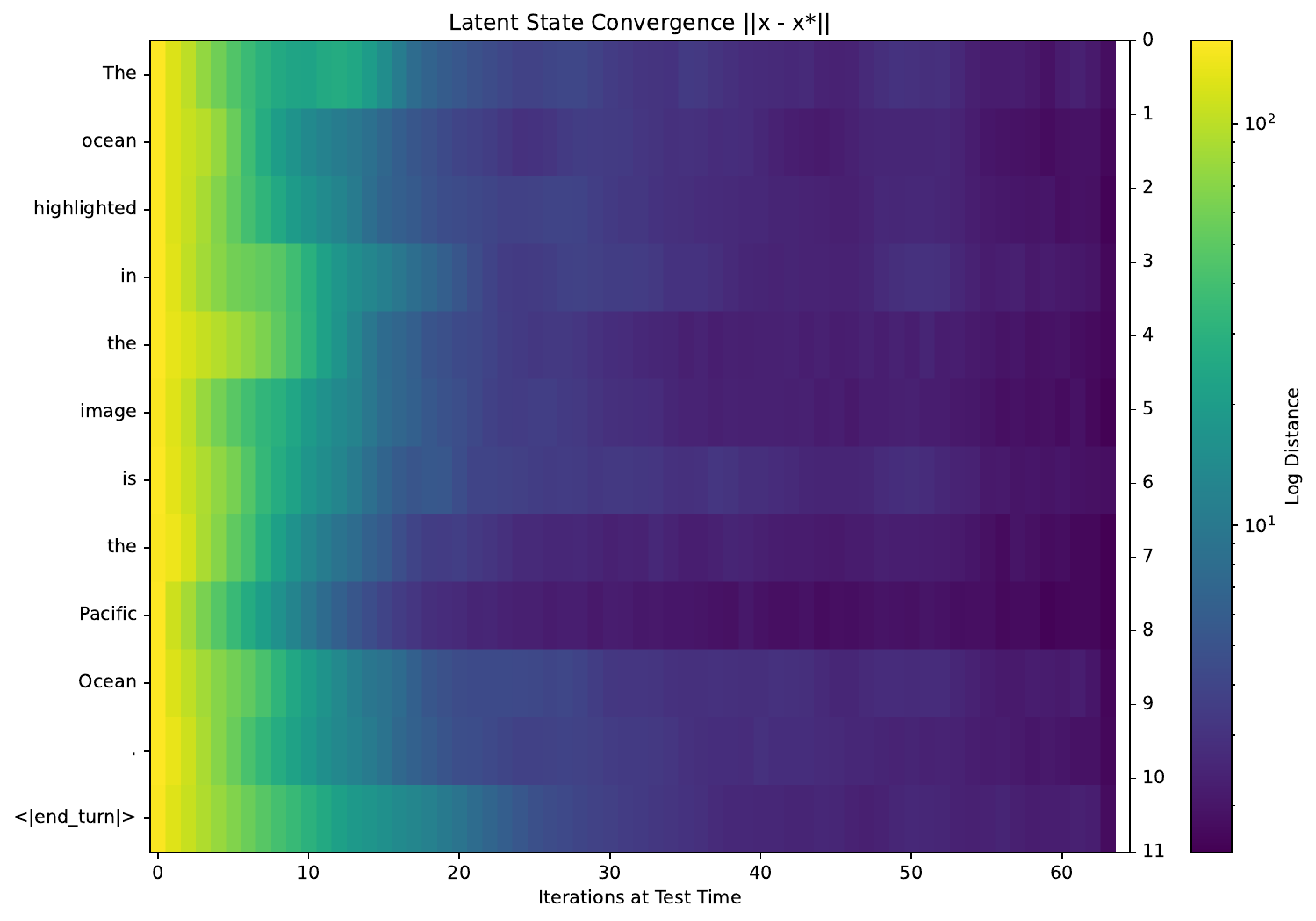}
  \caption{The visualization shows the evolution of latent states as a function of token position (vertical axis) and iteration depth (horizontal axis) in the model variant (r=32, w/o Hier,). Each cell represents the distance between a given iterate and its corresponding steady state, approximated at $r=32$.}
  \label{fig:baseline_heatmap}
\end{figure}

\begin{figure}
  \centering
  \includegraphics[width=0.8\linewidth]{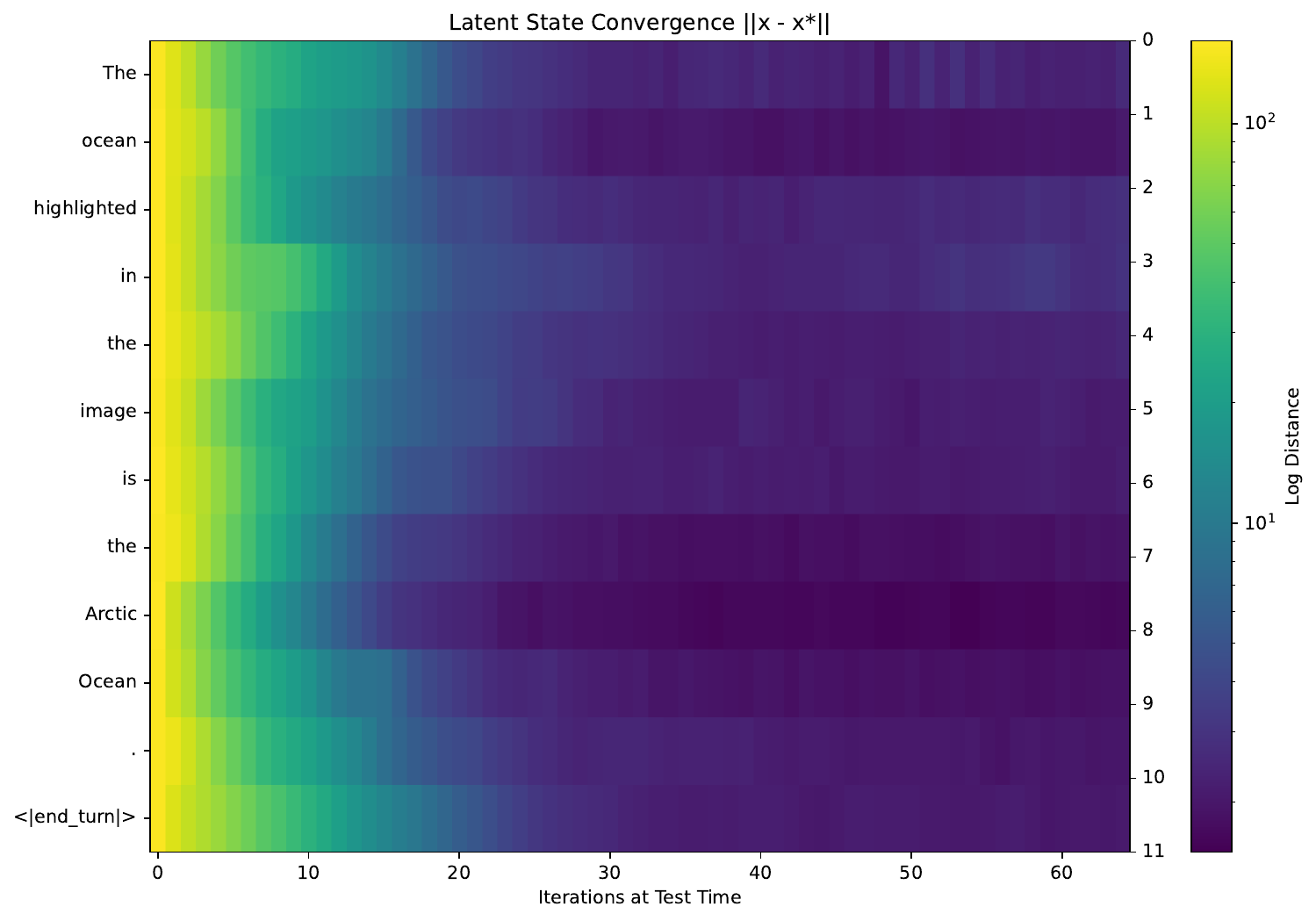}
  \caption{The visualization shows the evolution of latent states as a function of token position (vertical axis) and iteration depth (horizontal axis) in the \ourmodel. Each cell represents the distance between a given iterate and its corresponding steady state, approximated at $r=32$.}
  \label{fig:echo_heatmap}
\end{figure}

\begin{figure}
  \centering
  \includegraphics[width=0.75\linewidth]{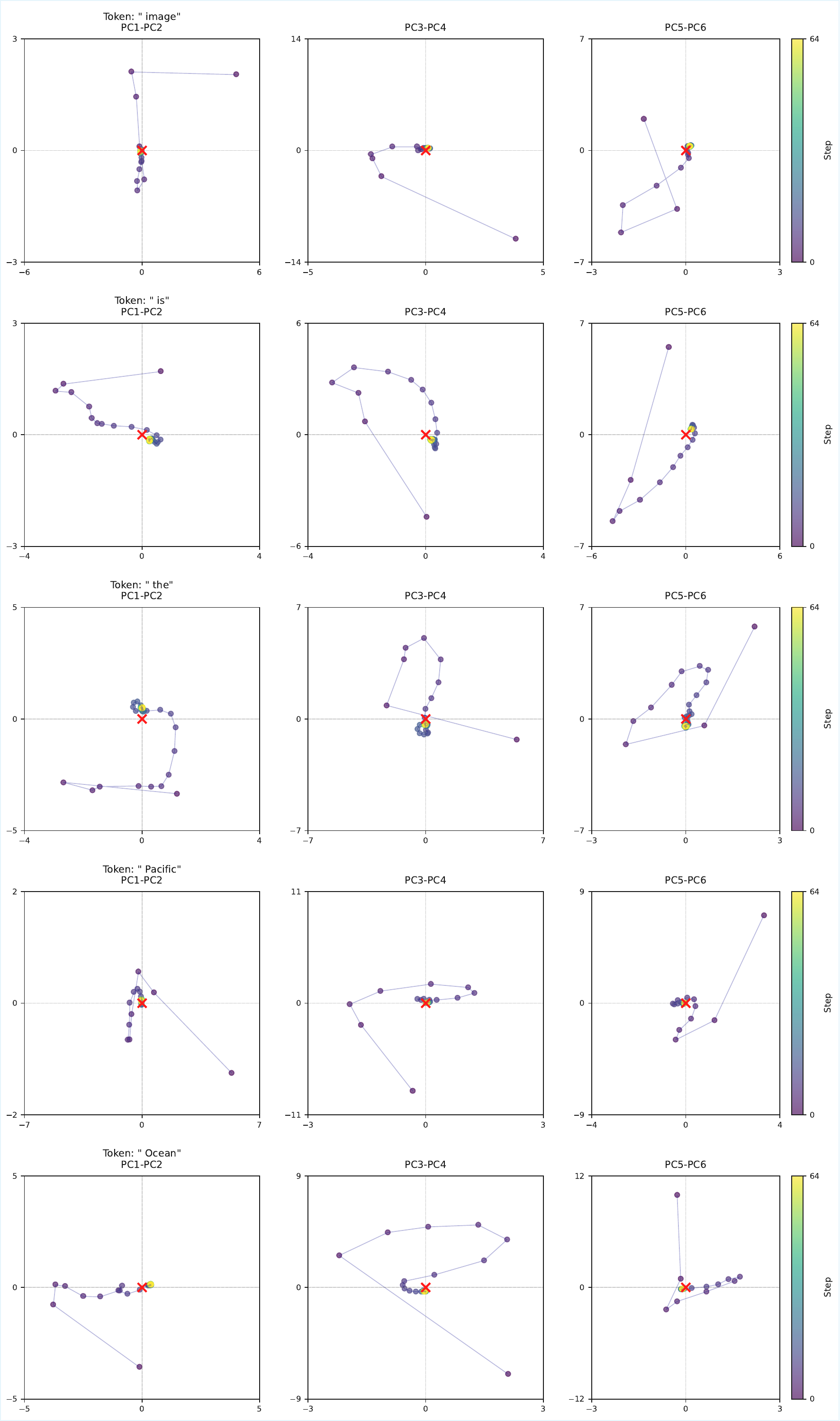}
  \caption{Visualization of hidden state trajectories in the model variant (r=32 w/o Hier.) during inference. }
  \label{fig:baseline_s}
\end{figure}

\begin{figure}
  \centering
  \includegraphics[width=0.75\linewidth]{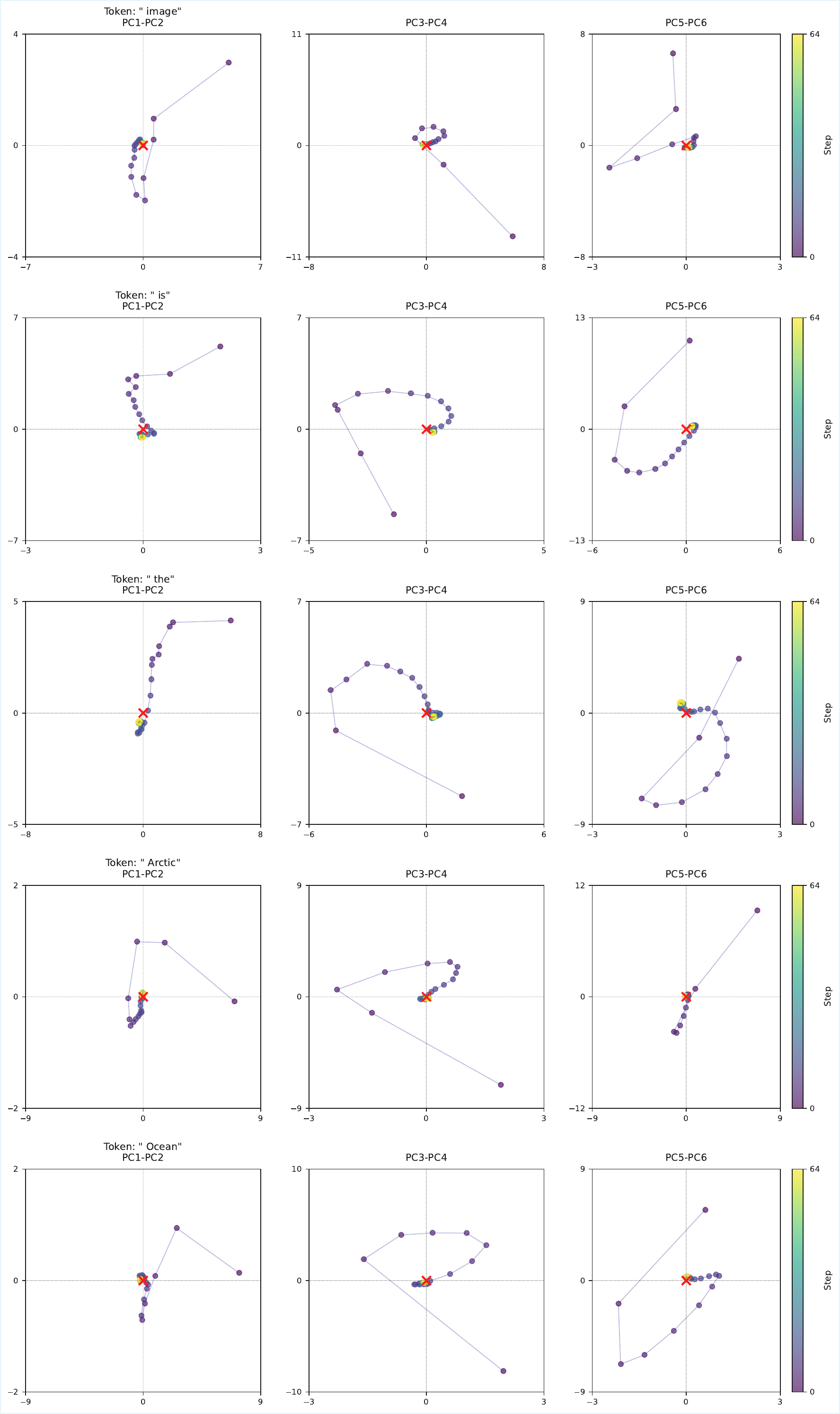}
  \caption{Visualization of hidden state trajectories in \ourmodel during inference.}
  \label{fig:echo_s}
\end{figure}

We visualize the internal mechanisms of our model using a specific case from ScienceQA, as shown in Figure \ref{fig:case}. To analyze convergence behavior, Figures \ref{fig:baseline_heatmap} and \ref{fig:echo_heatmap} compare the evolution of latent states toward a steady state, while Figures \ref{fig:baseline_s} and \ref{fig:echo_s} trace the specific trajectories of hidden states during inference for both the baseline  (r=32, w/o Hier,) and \ourmodel.

\section{Test-time scaling Results}
\begin{figure}
  \centering
  \includegraphics[width=\linewidth]{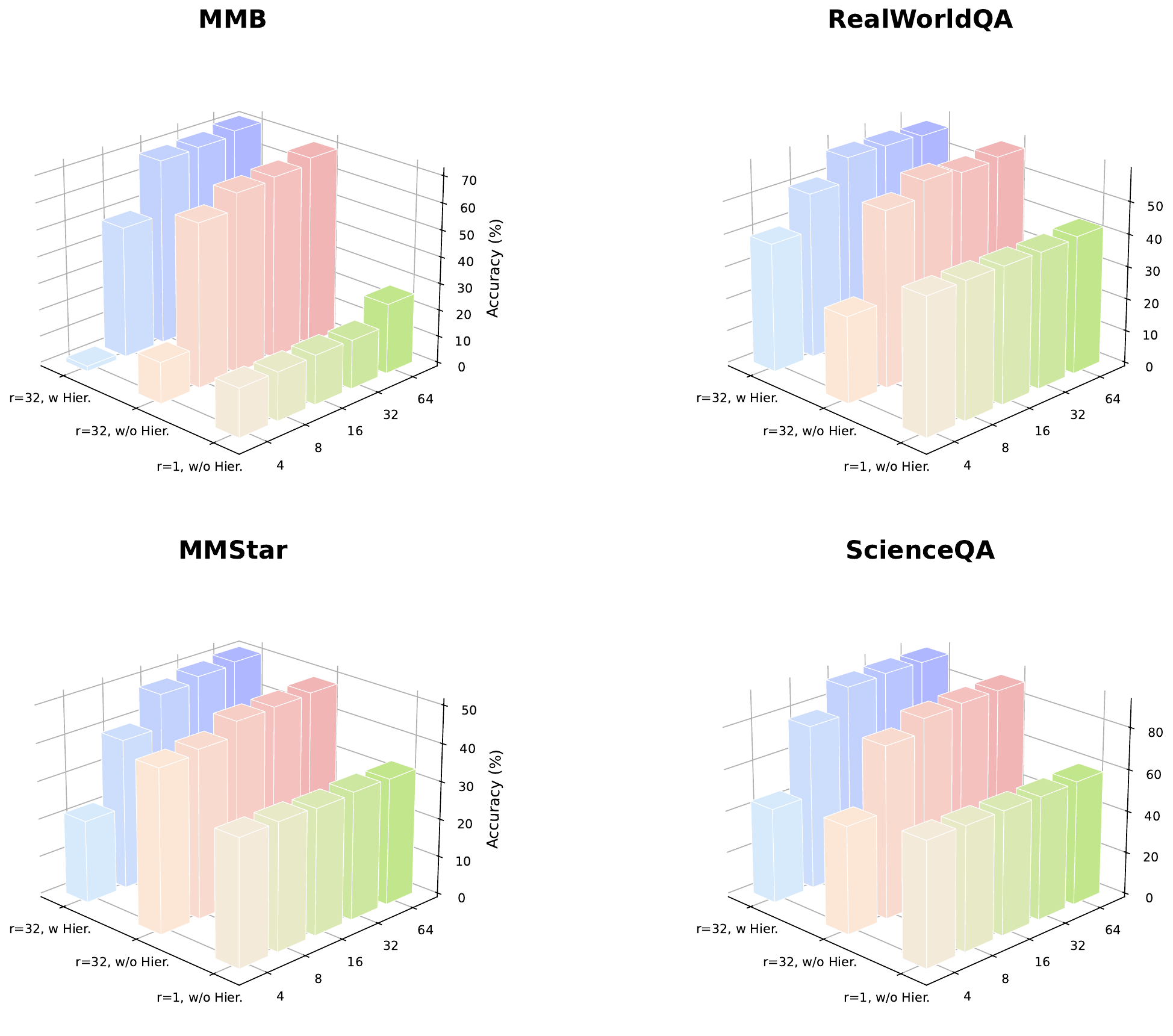}
  \caption{}
  \label{fig:bar_combined}
\end{figure}

The performance gains achieved through increased computational budget during inference are quantified in this section. Figure \ref{fig:bar_combined} visualizes the trends of model accuracy relative to scaling parameters at test-time.

\end{document}

%% file: sec/0_abstract.tex
\begin{abstract}
The advancement of multimodal large language models (MLLMs) has enabled impressive perception capabilities. However, their reasoning process often remains a "fast thinking" paradigm, reliant on end-to-end generation or explicit, language-centric chains of thought (CoT), which can be inefficient, verbose, and prone to hallucination. This work posits that robust reasoning should evolve within a latent space, integrating multimodal signals seamlessly. We propose multimodal latent reasoning via HIerarchical Visual cuEs injection (\emph{HIVE}), a novel framework that instills deliberate, "slow thinking" without depending on superficial textual rationales. Our method recursively extends transformer blocks, creating an internal loop for iterative reasoning refinement. It further injects hierarchical visual cues, from fine-grained regional details to coarse-grained scene-level semantics, into the model's latent representations, showing that this strategy remains effective in a loop-transformer reasoning framework. This enables the model to perform grounded, multi-step inference entirely in the aligned latent space. Extensive evaluations demonstrate that test-time scaling remains effective when incorporating vision knowledge, and that hierarchical visual cue injection can be effectively integrated into the loop-transformer framework for improved understanding of complex scenes.
\end{abstract}

%% file: sec/1_intro.tex
\section{Introduction}
The rapid evolution of large-scale pre-trained models \cite{brown2020language, jaredkaplan2020} has fundamentally transformed the landscape of artificial intelligence. Beginning with breakthroughs in natural language processing, models such as GPT-3  \cite{brown2020language} demonstrated unprecedented capabilities in understanding and generating human-like text. This progress soon expanded into the multimodal domain, where systems like GPT-4 \cite{DBLP:journals/corr/abs-2303-08774} and Qwen-VL~\cite{wang2024qwen2} have set new benchmarks by integrating and aligning information across vision, language, and beyond. These Multimodal Large Language Models (MLLMs)  \cite{DBLP:journals/corr/abs-2303-08774, liu2023visual} excel in multimodal tasks such as visual question answering, image captioning, and cross-modal retrieval. Their success marks a paradigm shift from unimodal intelligence toward more holistic, human-like understanding, enabling richer interactions and more robust applications in real-world scenarios.
\begin{figure}
  \centering
  \includegraphics[width=\columnwidth]{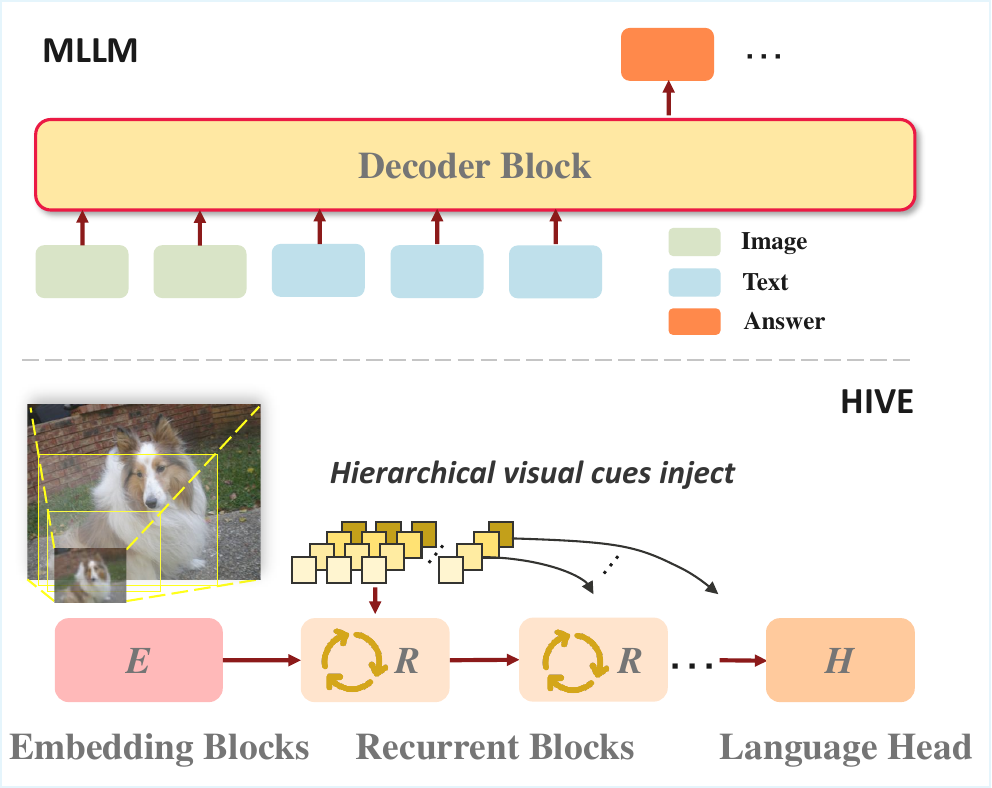}
  \caption{Visualization of traditional MLLMs, visual features extracted from a vision tower are projected into the language space and directly concatenated with text tokens. This combined sequence is then fed into a stack of transformer decoder blocks. \ourmodel is built upon Huginn, a recursive architecture that iteratively processes token representations through a unified set of layers to enhance feature depth. We have extended this by incorporating the visual modality and, for the first time, introducing hierarchical visual information into latent space reasoning.}
  \label{fig:cover}
\end{figure}

Building upon these advancements, the prevailing reasoning paradigm in most existing MLLMs can be characterized as a form of "System 1" or fast thinking which is a rapid, intuitive, and associative processing of multimodal inputs to generate direct, end-to-end responses. While effective for many pattern recognition and simple descriptive tasks, this single-step generation process often struggles with complex, compositional questions that demand deeper logical inference, sequential deliberation, or multi-faceted analysis. This limitation has spurred recent research aimed at instilling models with "System 2" or slow thinking capabilities, which involve explicit, structured, and often iterative reasoning steps. Representative efforts in this direction include LLaVA-CoT~\cite{xu2025llava}, which applies chain-of-thought prompting to visual questions; Vision-r1~\cite{huang2025vision}, which uses outcome-supervised rewards to incentivize faithful reasoning; and Mulberry~\cite{yao2024mulberry}, which employs collective tree search for deliberate planning. These works collectively highlight a critical shift towards more deliberate reasoning processes.

However, a common reliance on explicitly generated textual rationales as the primary scaffold for "slow thinking" may introduce inefficiencies and remain susceptible to the very language biases and hallucinatory pitfalls that deeper reasoning seeks to mitigate. This highlights the need for exploring more fundamental, latent, and modality-synchronized reasoning structures beyond surface-level linguistic chains. A series of works in LLMs focus on thinking in the latent space \cite{xinghaochen2025, shibohao2024, jindongli2025}, where the model performs computations entirely within its continuous hidden state. Recently, Heima \cite{xuanshen2025} extends the latent-space reasoning paradigm in LLMs to multimodal settings by introducing a set of Heima Encoder/Decoders. During the training stage, the encoder learns to compress CoT into predefined tokens. At inference time, Heima employs independently trained decoders to decode these abstract compressed tokens. However, the reasoning process is still driven by textual CoT supervision, rather than being grounded in or induced by visual representations. As a result, visual information is not truly integrated into the model's reasoning mechanism.

In this work, we propose \ourmodel{}, the first multimodal latent reasoning framework that enhances reasoning capability through recursive extension of transformer blocks and hierarchical injection of visual cues, as illustrated in Figure \ref{fig:cover}. The core of our framework is a looped transformer architecture, which performs test-time scaling via recurrent blocks, enabling iterative refinement of latent representations. We incorporate hierarchical visual information, from fine-grained regional details to coarse-grained scene-level semantics, into the latent space of the model. This design demonstrates that multi-scale visual cue injection remains effective when integrated with loop-transformer-based latent reasoning. Our contributions in this paper are summarized as follows:

\begin{itemize}
    \item We propose \ourmodel, the first MLLM that leverages loop transformers to enable recursive reasoning within the latent space, moving beyond the limitations of purely feed-forward architectures.
    \item We introduce hierarchical visual cue injection into the recurrent blocks, which allows the model to perform iterative reasoning guided by structural visual information.
    \item Extensive evaluations demonstrate that test-time scaling remains effective when incorporating vision knowledge, and that hierarchical visual cue injection works effectively within the loop-transformer framework on complex scene understanding.
\end{itemize}

%% file: sec/2_related_work.tex
\section{Related Work}
\begin{figure*}[t]
  \centering
  \includegraphics[width=\linewidth]{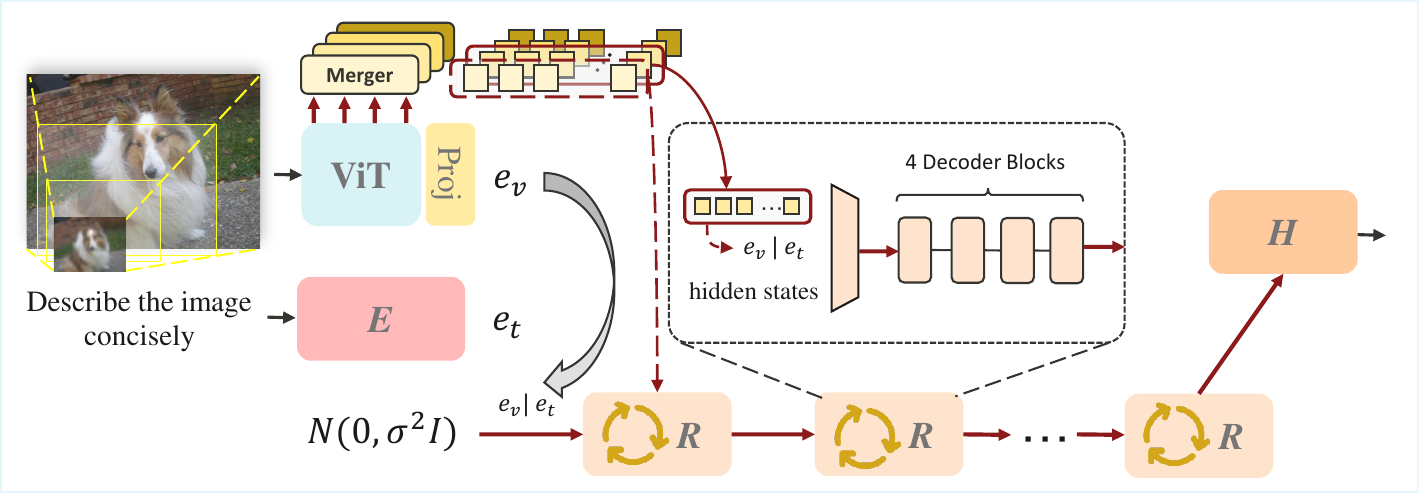}
  \caption{Our framework incorporates a pre-trained vision encoder, with a group of lightweight patch merger that maps visual features into the LLM embedding space. During multimodal alignment, the \texttt{[CLS]} token is removed. \protect\colorbox{P}{\rule{0pt}{5pt}\rule{10pt}{0pt}} \protect\colorbox{R}{\rule{0pt}{5pt}\rule{10pt}{0pt}} \protect\colorbox{C}{\rule{0pt}{5pt}\rule{10pt}{0pt}} represents \textit{Embedding}, \textit{Recurrent}, \textit{Head} blocks respectively.}
  \label{fig:framework}
\end{figure*}

\subsection{Thinking in Latent Space}
Recently, many models that perform reasoning process in the latent space have emerged. \cite{DBLP:journals/corr/abs-2507-06203}. Early work, such as Coconut \cite{shibohao2024}, each language reasoning step in CoT is gradually substituted by hidden states from model. Subsequent work like SoftCoT \cite{yigexu2025} generates “soft prompts” using a lightweight auxiliary model to serve as an initial CoT before formal reasoning, with frozen backbone LLM parameter, it prevents the catastrophic forgetting observed in Coconut and gain better results. While these approaches primarily focus on the language models, extending latent reasoning to the multimodal realm introduces unique challenges, such as the direct alignment of visual features with abstract logical steps. Consequently, recent research has begun to bridge this gap by integrating cross-modal inputs into the latent thinking process.

\begin{table}[t]
    \centering
    \caption{Comparison of Latent Space Reasoning Approaches. Coconut/Heima progressively abstract CoT text into learnable tokens without altering model architecture; Heima is the first to include multimodal input. Huginn/\ourmodel utilize a Loop Transformer architecture to reason via iterations, reducing dependency on CoT tokens. \ourmodel further introduces hierarchical visual features. }
    \label{tab:model_comparison}
    \resizebox{\columnwidth}{!}{
    \begin{tabular}{l c c c c}
        \toprule
        \textbf{Method} & \textbf{Visual} & \textbf{Text} & \textbf{Hierarchical} & \textbf{\makecell{CoT Data \\ Requirement}} \\
        \midrule
        \rowcolor{green!10}
        \multicolumn{5}{l}{\textit{\textbf{Training-induced Recurrence}}} \\
        Coconut & \xmark & \cmark & - & High \\
        Heima   & \cmark & \cmark & \xmark & High \\
        \midrule
        \rowcolor{blue!10}
        \multicolumn{5}{l}{\textit{\textbf{Loop Transformer Recurrence}}} \\
        Huginn  & \xmark & \cmark & - & Low \\
        \ourmodel (Ours) & \cmark & \cmark & \cmark & Low \\
        \bottomrule
    \end{tabular}
    }
\end{table}

Within the multimodal latent-space reasoning framework \cite{xuanshen2025, DBLP:journals/corr/abs-2508-12587}, Heima adopts a training-driven approach akin to Coconut, where textual CoT is progressively compressed into specialized "thinking tokens." While Heima achieves latent reasoning through this behavioral adaptation, it does not alter the fundamental model architecture. In contrast, loop transformer recurrence \cite{DBLP:conf/iclr/DehghaniGVUK19, DBLP:conf/iclr/MohtashamiPJ25, DBLP:conf/iclr/BaeFHJ0S25, DBLP:journals/tmlr/GaoZXSHL0LL25, DBLP:journals/corr/abs-2502-05171} introduces an explicit structural recurrence. This paradigm enables the iterative refinement of hidden states within a single forward pass by cycling through shared layers, effectively decoupling the depth of "thinking" from the physical parameter count. A representative implementation of this architectural philosophy is Huginn \cite{DBLP:journals/corr/abs-2502-05171}, which serves as the foundational backbone for our proposed framework.

Specifically, Huginn organizes decoder-only blocks into three functional groups, which we denote as the \textit{prelude (embedding)} blocks $E$, the \textit{core recurrent} block $R$, and the \textit{coda (language head)} $H$: $E$ embeds the input into the latent space and is executed once, $R$ carries out the iterative computation and is applied $r$ times with shared weights, and $H$ maps the final latent state to output probabilities. Only the parameters of $R$ are reused across iterations, so the effective depth of ``thinking'' is decoupled from the parameter count.

Huginn uses a BPE vocabulary of 65{,}536 tokens \cite{DBLP:conf/acl/SennrichHB16a}. The 3.5B-parameter model was pretrained on roughly 0.8T tokens and released without subsequent finetuning; approximately 1.5B parameters lie in the non-recurrent embedding blocks and language head, 1.5B in the core recurrent block, and 0.5B in the tied input embedding.

During training, the recurrent depth of each step is sampled from a heavy-tailed log-normal Poisson distribution $\Lambda$ with a targeted mean $\bar{r}+1$, and gradients are propagated only through the last $k$ iterations of the recurrent unit, i.e., truncated backpropagation through depth \cite{DBLP:conf/icassp/MikolovKBCK11} (Huginn fixes $k=8$). Peak activation memory and backward cost are therefore independent of $r$, which makes stochastic-depth training of such recursive structures tractable.

\subsection{Multimodal Large Language Models}
Early MLLMs focus on aligning visual and textual representations in a shared semantic space. CLIP \cite{DBLP:conf/icml/RadfordKHRGASAM21} demonstrates the effectiveness of large-scale contrastive pretraining for zero-shot transfer, while BLIP \cite{DBLP:conf/icml/0001LXH22} and its variants \cite{pmlr-v202-li23q, liu2025principled} extend this paradigm toward generative multimodal modeling by connecting pretrained vision encoders with large language models. These works lay the foundation for subsequent MLLMs.

Building upon large language models, LLaVA \cite{liu2023visual, liu2023improvedllava} introduces a projector-based connection scheme and instruction tuning to enable multimodal dialogue and reasoning. By directly mapping visual features into the language embedding space, LLaVA and its follow-ups achieve effective vision–language alignment with relatively low training complexity. However, due to the limited resolution of pretrained vision encoders, such approaches face challenges in tasks requiring fine-grained visual understanding.

Recent large-scale multimodal models emphasize unified training and improved visual feature utilization. LLaVA-OneVision-1.5 \cite{LLaVA-OneVision-1.5, xie2025region, DBLP:journals/tmlr/0080ZGZ00ZZL0L25} extends the LLaVA framework with a unified training pipeline to improve robustness across diverse visual tasks, while Qwen3-VL \cite{Qwen-VL, Qwen2-VL, Qwen2.5-VL, DBLP:journals/corr/abs-2511-21631}, introduces deepStack \cite{DBLP:conf/nips/MengYTDW0024} to hierarchically inject fine-grained visual features into early layers of large language models, strengthening vision–language interaction without increasing input tokens. Nevertheless, these models remain based on feed-forward Transformer architectures with statically encoded visual representations, limiting explicit iterative or latent-space reasoning.

Beyond architectural scaling, recent models highlight a shift in multimodal reasoning paradigms. The emergence of GPT-4o \cite{gpt4o, gpt4o-mini} demonstrates the feasibility of fully unified multimodal models, supporting vision, language, audio, and video within a single framework. Moreover, recent studies \cite{yao2024mulberry, xu2025llava} suggest that enhanced reasoning performance increasingly relies on transitioning from fast, single-pass inference toward slower, deliberative reasoning processes, enabling multi-step refinement and improved decision making. This paradigm shift motivates the exploration of multimodal models that explicitly support iterative and structured reasoning mechanisms.

%% file: sec/3_architecture.tex
\section{Method}
In this section, we introduce our motivation to inject hierarchical cues into the recurrent blocks and how we incorporate visual information into Huginn.

\subsection{Recurrent Visual-Language Backbone}
We denote the number of textual tokens as $n_t$, the number of visual tokens produced by the vision tower and projector as $n_v$, the length of the fused multimodal sequence as $n = n_t + n_v$, the hidden dimension of the model as $h$, and the vocabulary as $V$. Given $r$ as the recurrence count (the number of loop iterations) and $t$ as the recurrent step index, an input text sequence $x \in V^{n_t}$ and $\mathbf{X}_v$ denotes the sequence of flattened image patches, we process the textual and visual modalities separately. For the visual components, we denote the vision transformer and its associated projector as $\mathrm{ViT}(\cdot)$ and $\mathrm{Proj}(\cdot)$, respectively. The feature extraction and fusion process can be written as:
\begin{equation}
e = [e_v; e_t] = \mathrm{concat}(e_v, e_t),
\end{equation}
where
\begin{equation}
\begin{cases}
e_v = \mathrm{Proj}(\mathrm{ViT}(\mathbf{X}_v)) \in \mathbb{R}^{n_v \times h}, \\
e_t = E(x) \in \mathbb{R}^{n_t \times h}.
\end{cases}
\end{equation}
$e_v$ and $e_t$ represent the visual and textual embeddings, respectively, so that the fused sequence $e$ has length $n$. We denote $s_{t}$ as the hidden state after the $t$-th recurrent iteration, where $t = 0,1,\dots,r$. To stabilize the recurrent iterations, Huginn utilizes a random vector:
\begin{equation}
s_0 \sim \mathcal{N}(0, \sigma^2 I_{n \cdot h}).
\end{equation}
In the initial iteration, this vector is concatenated with the input embeddings along the channel dimension, which is subsequently mapped back to the original dimensionality by an adapter within the recurrent block $\mathcal{R}\text{-}Block$. In subsequent iterations, the hidden states derived from the preceding block are concatenated with the input embeddings. Let $\hat{e}_{v}^{(t)}$ be the fused visual cues:
\begin{equation}
s_{t+1}=\mathcal{R}\text{-}Block\left(e, \hat{e}_{v}^{(t)}; s_{t}\right).
\end{equation}

\subsection{Hierarchical Visual Injection}
To empower Huginn with the ability to perceive both structural details and high-level semantics, we move beyond the conventional practice of utilizing only the last layer of the vision encoder. Instead, we introduce a hierarchical visual injection strategy.

Specifically, we extract hidden states from a set of representative layers $\mathcal{L} = [6, 12, 18, 24]$. This selection is motivated by the inherent hierarchical nature of vision transformers:
\begin{itemize}
    \item Lower-level Layers (e.g., Layer 6) retain high-resolution spatial information and primitive visual patterns such as textures and edges, which are beneficial for grounding tasks.
    \item Intermediate and Higher-level Layers (e.g., Layer 12 to 24) gradually aggregate these primitives into complex semantic concepts and global context, providing the model with a holistic understanding of the scene.
\end{itemize}

To bridge the modality gap and align the dimensionalities, we employ a set of patch mergers inspired by Qwen3-VL: $\mathcal{M} = \{m_l\}_{l \in \mathcal{L}}$ \cite{DBLP:journals/corr/abs-2511-21631}. For each selected layer $l$, the visual features $h_v^l$ are projected as:
\begin{equation}
v_l = m_l(h_v^l), \quad l \in \{6, 12, 18, 24\},
\end{equation}
where $v_l \in \mathbb{R}^{n_v \times h}$ represents the projected visual cues ready for recurrent injection. By progressively injecting these features from fine-grained textures to coarse-grained semantics into the initial recurrent iterations, we provide the language backbone with a ``curriculum'' of visual understanding, stabilizing the hidden state transition during the early stages of reasoning.
\begin{equation}
\hat{e}_{v}^{(t)} =
\begin{cases}
v_{\mathcal{L}[t]} & \text{if } t < |\mathcal{L}|, \\
0 & \text{if } t \geq |\mathcal{L}|.
\end{cases}
\end{equation}

To enhance the robustness of the recurrent reasoning process, the recurrent depth $r$ is randomly sampled from a log-normal Poisson distribution during the training stage of Huginn. This stochasticity forces our model to decouple the visual-language fusion from a fixed step count.

We introduce an adaptive injection schedule. The core challenge lies in aligning the $r$ recurrent iterations with the $|\mathcal{L}|$ available visual tiers. We define the injection according to the sampled depth $r$ as follows:

\begin{enumerate}
    \item \textbf{Case I: Sufficient Iterations} ($r \geq |\mathcal{L}|$). The model injects one hierarchical visual tier per recurrent step during the first $|\mathcal{L}|$ iterations, and uses the remaining iterations for pure latent refinement.
    \item \textbf{Case II: Constrained Iterations} ($r < |\mathcal{L}|$). Only the first $r$ visual tiers are injected, one per step, since the recurrence terminates before all hierarchical cues are consumed.
\end{enumerate}

\lstset{
    language=Python,
    basicstyle=\ttfamily\scriptsize\bfseries,
    keywordstyle=\color{blue}\bfseries,
    commentstyle=\color{green!50!black},
    stringstyle=\color{orange},
    numbers=left,
    numberstyle=\tiny,
    stepnumber=1,
    numbersep=5pt,
    backgroundcolor=\color{gray!5},
    breaklines=true,
    breakatwhitespace=true,
    showstringspaces=false,
    tabsize=4,
    aboveskip=10pt,
    belowskip=10pt,
    lineskip=2pt,
    columns=flexible,
}

\begin{lstlisting}
def core_block_forward(x_in, embd):
    ... # Model expand recurrent blocks here.
    return x_out

def iterate_forward(x, embeds, vis_features):
    n_no_grad, n_grad = random_sampler()

    def get_input(i):
        if i < len(vis_features):
            # Vision features are injected into embeds
            return func(embeds, vis_features[i])
        else:
            return embeds

    with torch.no_grad():
        for i in range(n_no_grad):
            core_block_forward(x, get_input(i))

    for i in range(n_no_grad, n_no_grad + n_grad):
        core_block_forward(x, get_input(i))
    ...
\end{lstlisting}

Finally, after $r$ recurrent iterations, the model decodes the hidden state $s_r$ to produce the output probabilities:
\begin{equation}
p = H(s_r).
\end{equation}

\subsection{Training Objective}
Given an input text sequence $x$ and a set of hierarchical visual features $\mathbf{V}_{hier} = \{v_{(1)}, v_{(2)}, \dots, v_{(|\mathcal{L}|)}\}$ extracted from multiple encoder layers, the training loss is defined as:
\begin{equation}
\mathcal{L}(\theta) = \mathbb{E}_{(x, \mathbf{V}_{hier}) \in \mathcal{X}}\, \mathbb{E}_{r \sim \Lambda} \left[ \mathcal{L}_{\text{CE}} \left( f_\theta(x, \mathbf{V}_{hier}, r),\, x' \right) \right],
\end{equation}
where $\theta$ denotes the trainable parameters, and $f_\theta(x, \mathbf{V}_{hier}, r)$ represents the model output at recurrence depth $r$. The hierarchical features $\mathbf{V}_{hier}$ are selectively injected into the early recurrent iterations. This ensures that the model progressively refines its latent representations by anchoring them to multi-scale visual cues. The recurrence depth $r$ is sampled from a log-normal Poisson distribution $\Lambda$ with a targeted mean $\bar{r}+1$. This stochastic supervision forces the model to maintain semantic consistency across varied computational paths, facilitating the adaptive early-exit mechanism during inference.

%% file: sec/4_experiments.tex
\section{Experiments}
\subsection{Training Configuration}
Following the established methodology of MLLMs like LLaVA-NeXT \cite{liu2024llavanext}, we decouple the training of \ourmodel into a three-stage pipeline.

\begin{table}[t]
\centering
\caption{Comparison of Various Backbone LMMs, including Gemma3 \cite{DBLP:journals/corr/abs-2503-19786}, LLaMA2 \cite{DBLP:journals/corr/abs-2307-09288}, Phi-3-mini \cite{DBLP:journals/corr/abs-2404-14219} and Huginn \cite{DBLP:journals/corr/abs-2502-05171}. We detail model sizes, training token counts, and benchmark performances. The results indicate that the inherent language capabilities of our backbone LLM are constrained, which may, in turn, impact the overall performance of the resulting MLLMs. The missing entries are unavailable results.}
\begin{tblr}{
  width = \linewidth,
  colspec = {ccccc},
  colsep = 2pt,
  cells = {c},
  hline{1,7} = {-}{0.08em},
  hline{2,4} = {-}{0.05em},
}
             & Gemma3 & LLaMA2& Phi-3-mini & Huginn \\
Param        & 4B     & 7B     & 3.8B       & 3.5B   \\
Train Tokens & 4T     & 2.0T   & 3.3T       & 0.8T   \\
GSM8K        & 38.4   & -      & -          & 28.20  \\
\makecell{GSM8K-CoT \\ (8-shot)}    & -      & -      & 82.5       & 34.57  \\
HumanEval    & 36.0   & 29.9   & 58.5       & 23.17  
\end{tblr}
\label{tab:backbone}
\end{table}

To capture fine-grained visual details across varying aspect ratios, we employ InternViT-300M-448px-V2.5 as our visual backbone. Unlike static encoders that resize images to a fixed square, our model leverages a dynamic high-resolution strategy. In our implementation, we set the maximum number of image tiles to a relatively conservative value to align with our available computational budget. Details are shown in Table \ref{tab:training_stages}.

\paragraph{Stage 1.} To pre-align the vision-language modality, we only train the projector as well as the patch mergers in this stage, using the LCS-558K dataset \cite{liu2023visual}. 

\begin{table}
\small
\setlength{\tabcolsep}{4pt} 
\caption{Training configurations across different stages.}
\label{tab:training_stages}
\begin{tabular}{lccc}
\toprule
\textbf{Parameters} & \textbf{Stage 1} & \textbf{Stage 2} & \textbf{Stage 3} \\ 
\midrule
Learning Rate      & $1 \times 10^{-3}$ & $1 \times 10^{-5}$ & $1 \times 10^{-6}$ \\
Max Dynamic Patches (\#)    & 4                  & 2                  & 2                  \\
Max Tokens              & 1536               & 2048               & 2048               \\ 
\bottomrule
\end{tabular}
\end{table}

\begin{table*}[t]
\centering
\caption{\textbf{Main performance comparison with MLLMs.} We report the LLM backbone, model size, and training data scales (4T* denotes 4T tokens, otherwise samples). Note that backbones are detailed in Table \ref{tab:backbone}. MMB: MMBench (En) dev split; SQA$_{\text{img}}$: ScienceQA-Img; SEED$_{\text{img}}$: SEED-Bench-Img; RWQA: RealWorldQA; TVQA: TextVQA; CQA: ChartQA; DVQA: DocVQA; POPE; Missing entries indicate unavailable results.}
\begin{tblr}{
  width = \linewidth,
  colspec = {ccccccc},
  column{3-8} = {c},
  cell{1}{1} = {c=2}{c, m}, 
  cell{2}{1} = {c=2}{c}, 
  cell{3}{1} = {c=2}{c},
  cell{4}{1} = {c=2}{c},
  cell{5}{1} = {r=4}{c, m}, 
  cell{9}{1} = {r=3}{c, m}, 
  cell{12}{1} = {r=1}{c, m}, 
  hline{1,13} = {0.08em},    
  hline{2,5,9,12} = {1-8}{0.03em},
}
& & Gemma3-4B-PT & MobileVLM V2 7B & Bunny-v1.1-4B & Imp-4B & Emu3 & \ourmodel \\
Backbone & & Gemma3 & Vicuna-7B-v1.5 & Phi-3-3.8B & Phi-3-3.8B & - & Huginn \\
Param. & & 4B & 7B & 4B & 4B & 8B & 4B \\
Train Data & & 4T* & 3.6M & 2.7M & 1.6M & - & 6.5M \\
General VQA & MMB & - & 69.2 & 74.2 & 73.3 & 58.5 & 69.6 \\
 & SQA$_{\text{img}}$ & - & 74.8 & 78.3 & 78.3 & 89.2 & 91.6 \\
 & SEED$_{\text{img}}$ & - & - & 72.5 & - & 68.2 & 70.5 \\
 & RWQA & 45.5 & - & - & - & 57.4 & 57.9 \\
OCR Chart & TVQA$_{\text{val}}$ & 58.9 & 62.3 & - & 60.2 & 64.7 & 57.5 \\
 & CQA$_{\text{test}}$ & 63.6 & - & - & - & 68.6 & 67.0 \\
 & DVQA$_{\text{val}}$ & 72.8 & - & - & - & 76.3 & 73.2 \\
Others & POPE & - & 85.3 & 87.2 & 86.9 & 85.2 & 87.6 \\
\end{tblr}
\label{tab:main_results}
\end{table*}

\paragraph{Stage 2.} To enhance image-text alignment, we use the subset of the EMOVA alignment dataset \cite{kaichen2025}. General visual-language pre-training is sourced from ShareGPT4V \cite{DBLP:conf/eccv/ChenLDZHWZL24}, ALLaVA \cite{DBLP:journals/corr/abs-2402-11684} (English and translated Chinese), and ShareGPT-4o \cite{cui2025comprehensive}, while OCR-related capabilities are supported by SynthDog \cite{DBLP:conf/eccv/KimHYNPYHYHP22}, MMC-Alignment \cite{DBLP:conf/naacl/LiuWYCSCYY24}, K12 Printing, and the UReader Text Reading subset \cite{DBLP:conf/emnlp/YeHXYYXLT0ZJHLH23}. We also integrate the text-only corpus from Magpie Pro \cite{DBLP:conf/iclr/XuJNDP0L25} into our multi-modal training pipeline to maintain strong linguistic proficiency. 
\paragraph{Stage 3.} Our Stage 3 training data comprises 3.4M samples from the EMOVA-SFT subset, together with a collection of high-quality open-source visual instruction datasets, including ShareGPT4V \cite{DBLP:conf/eccv/ChenLDZHWZL24}, InternVL \cite{chen2024internvl}, Meteor \cite{DBLP:conf/nips/LeeKPR24}, Idefics-2 \cite{DBLP:conf/nips/LaurenconTCS24}, Cambrian \cite{tong2024cambrian}, and LLaVA-OneVision \cite{DBLP:journals/tmlr/0080ZGZ00ZZL0L25}. See Fig. \ref{fig:data_stat} for more details.

\subsection{Setup}
\paragraph{Tokenization.} To bridge the modalities, we introduce three dedicated tokens: \texttt{<|image\_start|>}, \texttt{<|image|>}, and \texttt{<|image\_end|>}. Specifically, the \texttt{<|image|>} token serves as a placeholder and is substituted by the visual features projected from the vision encoder into the language embedding space.

\paragraph{Hyperparameters.} We train the model with a weight decay of $10^{-3}$. We adhere to the optimization configuration of the original Huginn, employing the AdamW optimizer with $\beta_{1} = 0.9$ and $\beta_{2} = 0.95$. The learning rate is set with a cosine decay scheduler. Other detailed configurations are summarized in Table~\ref{tab:training_stages}.

\paragraph{Evaluation.}

We evaluated the effectiveness of our approach across several challenging benchmarks via LMMs-Eval \cite{zhang2024lmmsevalrealitycheckevaluation}, including MMStar \cite{DBLP:conf/nips/ChenLDZZCDWQLZ24}, MMBench \cite{DBLP:conf/eccv/LiuDZLZZYWHLCL24}, ScienceQA \cite{DBLP:conf/nips/LuMX0CZTCK22}, SEED-Bench \cite{DBLP:journals/corr/abs-2307-16125}, and RealWorldQA for general visual question answering capability. For OCR \& Chart, we utilize ChartQA \cite{DBLP:conf/acl/MasryLTJH22}, TextVQA \cite{singh2019towards}, and DocVQA \cite{DBLP:conf/wacv/MathewKJ21}. In addition, we use MathVista \cite{DBLP:conf/iclr/LuBX0LH0CG024} for math and reasoning evaluation. POPE \cite{li2023evaluating} and GQA \cite{DBLP:conf/cvpr/HudsonM19} are adopted to assess model capabilities in hallucination-prone scenarios and complex visual reasoning challenges.

\subsection{Benchmark Results}
We developed three models for a comparative study: a baseline model trained without recurrence, a model trained with a mean recurrence of 32, and a third model that incorporates hierarchical visual information with the same recurrence level. To assess performance, we benchmarked these models against several open-source models, including Gemma-3-4B-PT \cite{DBLP:journals/corr/abs-2503-19786}, MobileVLM V2 7B \cite{DBLP:journals/corr/abs-2402-03766}, Bunny-v1.1-4B \cite{DBLP:journals/corr/abs-2402-11530}, Imp-4B \cite{DBLP:journals/tmm/ShaoYYOZGWKD25}, and Emu3 \cite{DBLP:journals/corr/abs-2409-18869}. The main results are shown in Table \ref{tab:main_results}. \ourmodel is evaluated using $r = 32$.

\begin{table}
\small
\setlength{\tabcolsep}{6pt} 
\centering
\caption{Results on benchmarks for three variants. Bold text indicates the best performance among these models. MMB: MMBench (En) dev split; SQA$_{\text{img}}$: ScienceQA-Img; SEED$_{\text{img}}$: SEED-Bench-Img; RWQA: RealWorldQA; TVQA: TextVQA; DVQA: DocVQA; MathV: MathVista; POPE}
\begin{tabular}{l|c|c|c} 
\toprule
\textbf{Benchmark} & \makecell[c]{\textbf{Baseline} \\ \textbf{(r=1)}} & \makecell[c]{\textbf{Ours w/o Hier} \\ \textbf{(r=32)}} & \makecell[c]{\textbf{Ours w/ Hier} \\ \textbf{(r=32)}} \\ 
\midrule
MMStar & 33.28 & 48.44 & \textbf{49.79} \\
SEED$_{\text{img}}$ & 42.37 & 70.46 & \textbf{70.48} \\
MMB$_{\text{dev}}$ & 21.74 & 68.04 & \textbf{69.59} \\
RWQA & 41.44 & 57.52 & \textbf{57.91} \\
SQA$_{\text{img}}$ & 60.09 & 89.39 & \textbf{91.57} \\
\midrule
\textbf{Avg.} & 39.78 & 66.77 & \textbf{67.87} \\ 
\midrule

DVQA$_{\text{val}}$ & 24.04 & \textbf{73.72} & 73.20 \\
TVQA$_{\text{val}}$ & 30.56 & \textbf{57.66} & 57.54 \\
\midrule
\textbf{Avg.} & 27.3 & \textbf{65.69} & 65.37 \\ 
\midrule
MathV$_{\text{mini}}$ & 24.5 & \textbf{35.0} & 34.7 \\

POPE & 74.84 & 87.02 & \textbf{87.61} \\
GQA & 44.8 & 57.71 & \textbf{57.89} \\
\midrule
\textbf{Avg.} & 48.05 & 59.91 & \textbf{60.07}\\ 
\bottomrule 
\end{tabular}
\label{tab:simplified_results}
\end{table}

Based on the results, \ourmodel demonstrates a competitive edge in parameter efficiency and specialized visual reasoning. Despite its compact 4B architecture, the model achieves 91.6 on ScienceQA-Img, notably outperforming the larger 8B Emu3 and the 7B MobileVLM V2. This indicates that \ourmodel is particularly effective at handling complex, knowledge-based visual tasks. Furthermore, it achieves the highest reliability in the POPE benchmark (87.6), suggesting a robust capability to mitigate object hallucination compared to its peers.

The model also exhibits impressive data efficiency. While trained on 6.5M samples, \ourmodel consistently outperforms or matches models like Gemma3-4B-PT, which benefit from a much larger 4T token pre-training scale, across benchmarks such as RealWorldQA and DocVQA. Overall, \ourmodel strikes a balance between model size and performance. This is particularly notable because it manages to overcome the inherent limitations of its relatively lightweight Huginn backbone to achieve results that rival established baselines.

There remains substantial space for performance optimization. We recognize that the current model can be further elevated through finer hyperparameter tuning and more sophisticated dynamic resolution configurations, which could better capture the intricate spatial details required for advanced OCR and document understanding tasks.

\begin{figure}[t]
  \centering
  \includegraphics[width=\columnwidth]{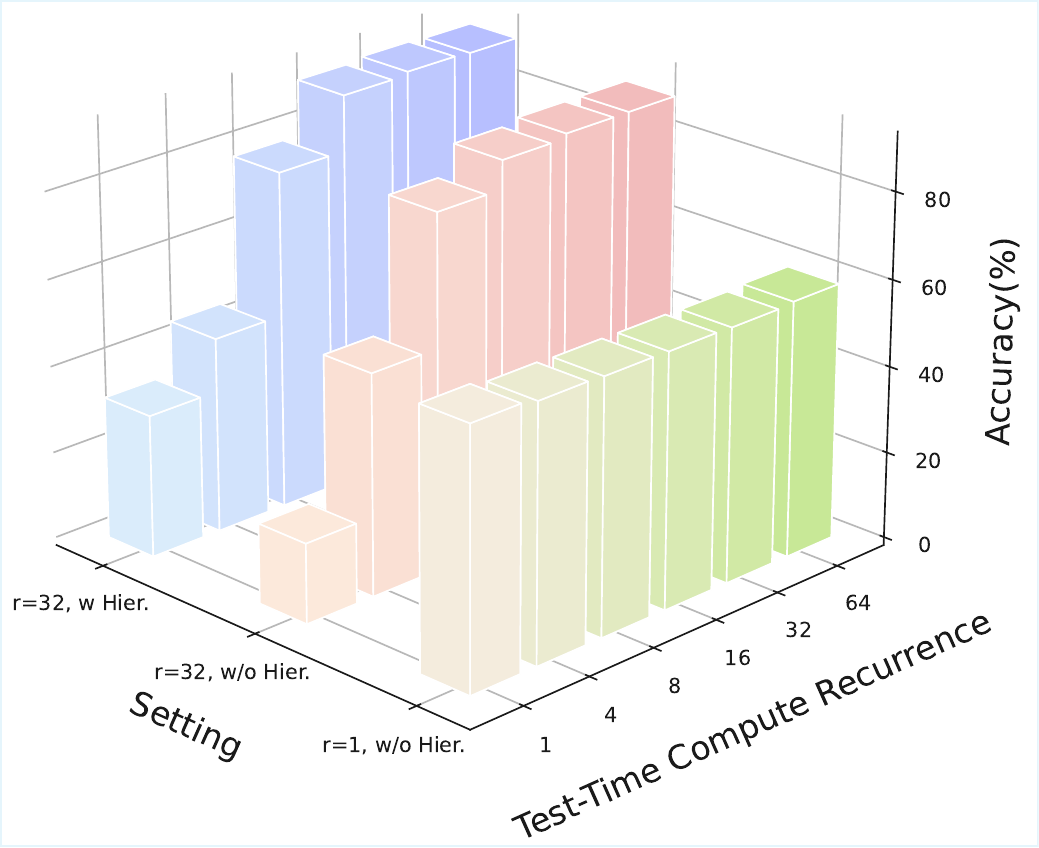}
  \caption{Building upon Huginn, we integrate a Vision Transformer (ViT) and propose a hierarchical latent-space reasoning framework. Specifically, we argue that latent-space reasoning with visual information should be hierarchical rather than merely iterative. The figure shows our comparison results on ScienceQA$_{\text{img}}$.}
  \label{fig:bar}
\end{figure}

\begin{figure}
  \centering
  \includegraphics[width=\linewidth]{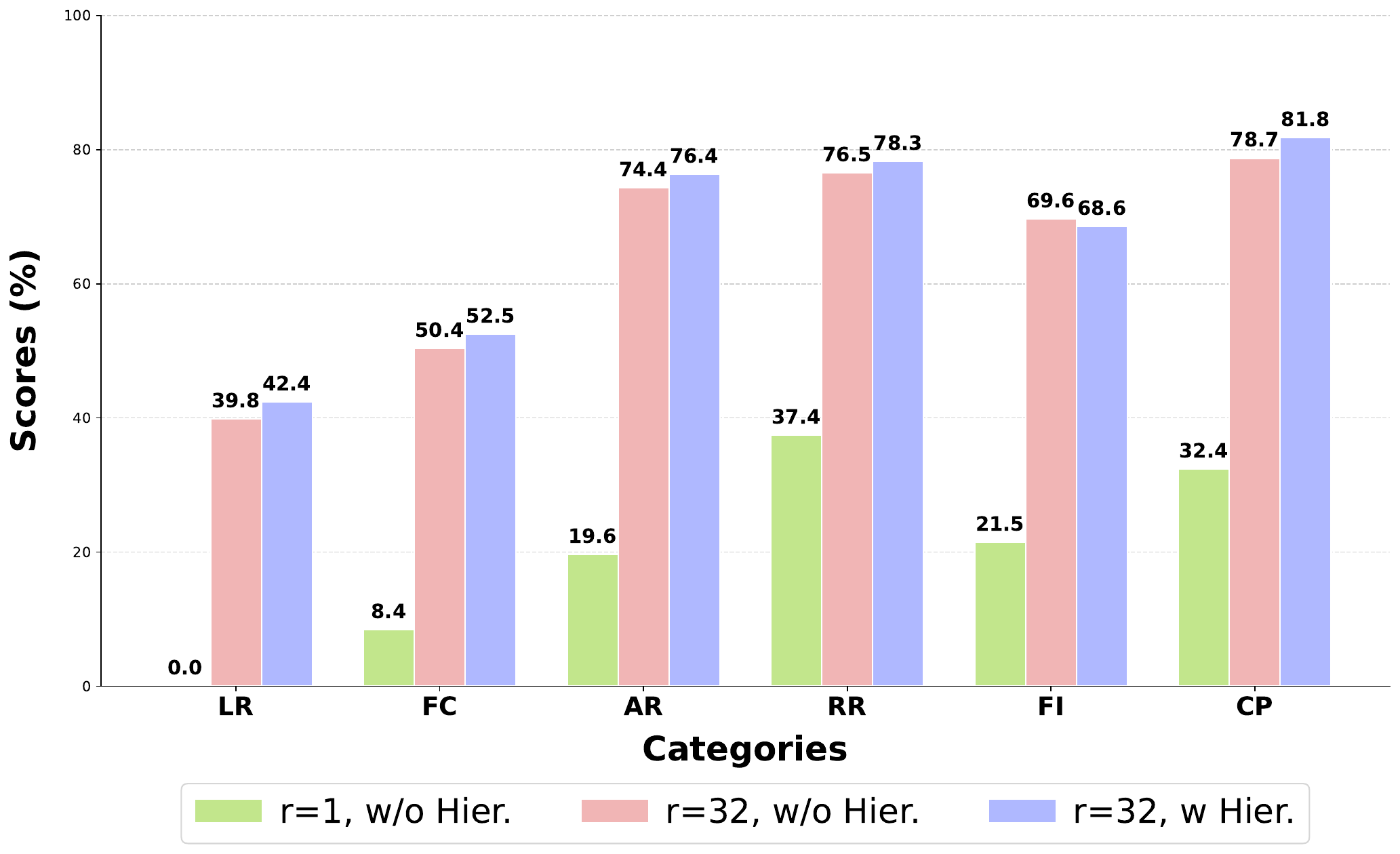}
  \caption{MMBench detailed results. LR denotes logic reasoning. FC denotes fine-grained perception (cross-instance). AR denotes attribute reasoning. RR denotes relation reasoning. FI denotes fine-grained perception (instance-level). CP denotes coarse perception.}
  \label{fig:mmb_bar}
\end{figure}

\begin{figure*}
  \centering
  
  \includegraphics[width=\linewidth]{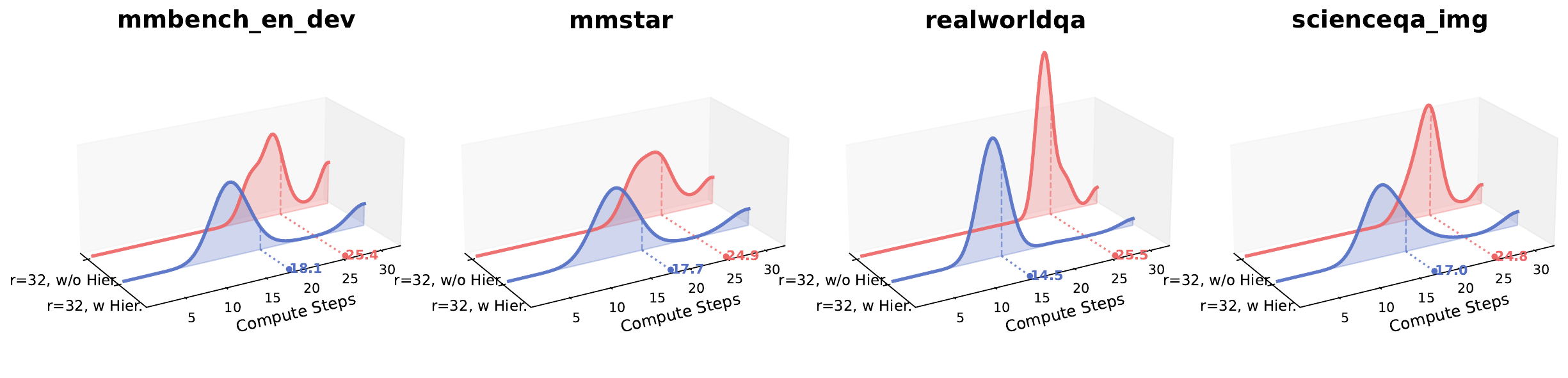}
  \caption{\textbf{Distribution of inference steps for the first token generation across multiple-choice benchmarks.} We evaluate the impact of hierarchical cue injection on the recurrent steps. The results demonstrate that incorporating these cues causes a distinct leftward shift in the distribution, indicating a reduction in computation during inference.}
  \label{fig:compute_steps}
\end{figure*}

\paragraph{Recurrence improves the performance.} Figure \ref{fig:bar} illustrates the performance scaling of three model variants across varying recurrence steps $r$: (1) a non-recurrent baseline (trained with $r=1$), (2) a recurrent variant without hierarchical cues ($\bar{r}=32, \text{w/o Hier.}$), and (3) our full recurrent model with hierarchical visual cues ($\bar{r}=32, \text{w/ Hier.}$). The empirical results yield several key insights:
\begin{itemize}
    \item Iterative Refinement Gains: While the non-recurrent baseline remains stagnant at a low performance level (averaging ~59.0\% across all steps), both recurrent variants exhibit a dramatic upward trajectory as $r$ increases. For instance, the hierarchical model climbs from 32.82\% at $r=1$ to a peak of 91.57\% at $r=32$, validating that iterative recurrence allows the model to progressively refine its internal representations.
    \item Impact of Hierarchical Cues: The incorporation of hierarchical cues is associated with modest performance gains in this setting. At the recurrence depth of $r=32$, the full model reaches 91.57\%, compared with 89.39\% for the ``w/o Hier.'' variant.
    \item Performance Saturation: We observe a clear "diminishing returns" effect beyond $r=32$. The performance gains for the hierarchical model plateau, moving from 91.57\% ($r=32$) to a slight fluctuation at 91.27\% ($r=64$). This convergence indicates that the model's representational capacity saturates at this depth, where additional computational steps no longer yield meaningful accuracy improvements.
\end{itemize}

\paragraph{Hierarchical cues help understanding.} To further examine the role of hierarchical visual cues, we report fine-grained results across six core dimensions in Figure \ref{fig:mmb_bar}. Compared with the recurrent baseline ($\bar{r}=32, \text{w/o Hier.}$), the hierarchical recurrent variant ($\bar{r}=32, \text{w/ Hier.}$) shows generally positive trends in several categories. In particular, \ourmodel yields moderate gains in Logic Reasoning (LR, $+2.54\%$), Attribute Reasoning (AR, $+1.99\%$), Relation Reasoning (RR, $+1.74\%$), and Coarse Perception (CP, $+3.04\%$), suggesting that hierarchical visual priors can be incorporated effectively into the recurrent framework. Although the differences are limited in instance-level perception (FI), the overall results indicate that hierarchical cues are compatible with loop-based latent reasoning and can provide additional support in complex visual understanding.

\subsection{Adaptive Compute}
To optimize the efficiency-performance trade-off, Huginn has implemented an adaptive computation mechanism that dynamically adjusts the number of recurrence iterations during inference. This optional mechanism lets the model determine the termination of recurrence based on the convergence of hidden states. A relative change metric is defined as follows:
\begin{equation*}
\text{norm\_diff} = \frac{\| \mathbf{h}_{t} - \mathbf{h}_{t-1} \|_2}{\| \mathbf{h}_{t} \|_2}.
\end{equation*}

To further enhance inference efficiency, Huginn adopts a specialized KV-cache management scheme with a periodic retrieval strategy. For the $i$-th token during the $r$-th recurrence step, the \textit{latest-m4} mechanism retrieves the KV-cache from the most recent valid step $j$ that aligns with the current block's functional cycle. Specifically, the \textit{latest-m4} lookup mechanism determines the retrieved cache index $j^*$ as:$$j^* = 
\begin{cases} 
\max \{ j \mid j \le r, j \equiv_4 r, \mathcal{I}_{j,i}=1 \} & r \ge 2, \\
r & r < 2,
\end{cases}$$where $\mathcal{I}_{j,i} \in \{0, 1\}$ denotes the validity of the cache at step $j$ for token $i$, and $\equiv_4$ denotes congruence modulo 4. This periodic reuse of cache states maintains temporal consistency while significantly reducing redundant computations.

This optimized caching framework provides the necessary infrastructure for dynamic inference. To quantify the computational effect, we analyze the average recurrence steps required for the first token under the adaptive early-exit setting (max $r=32$). As shown in Figure \ref{fig:compute_steps}, incorporating hierarchical cues is associated with faster convergence of hidden states across MMBench, MMStar, RealWorldQA, and ScienceQA$_{\text{img}}$. Concretely, the mean reasoning steps decrease from 25.4 to 18.1 on MMBench, 24.9 to 17.7 on MMStar, and 24.8 to 17.0 on ScienceQA$_{\text{img}}$. On RealWorldQA, the average computation depth decreases from 25.5 to 14.5. This leftward shift in the step distribution suggests that hierarchical visual cues can provide useful multi-scale information that helps the model meet the exit criterion earlier in some cases. Overall, these results indicate that hierarchical cue injection is compatible with reducing the number of recurrence steps required under adaptive computation.

%% file: sec/5_conclusion.tex
\section{Conclusion}
In this work, we introduced \ourmodel, a novel MLLM that pioneers the use of loop-based Transformer architectures for latent-space reasoning. By progressively leveraging hierarchical visual features through iterative recurrence, \ourmodel demonstrates that complex multimodal tasks can be refined within a fixed-parameter recurrent framework. Our experiments reveal that the integration of hierarchical cues is naturally suited to loop-based architectures and can improve reasoning efficiency. Under an adaptive computation setting, the hierarchical mechanism facilitates faster convergence of latent states.

Looking ahead, we aim to enhance OCR \& Chart performance via dynamic resolution strategies and investigate various layer-selection schemes. A primary focus is internalizing explicit CoT within recurrent loops. Central to this effort is the challenge of implementing early-exit mechanisms that reduce computational overhead while maintaining accuracy. Furthermore, we intend to explore the scalability of this recurrent approach to more diverse modal inputs. This research provides a practical path toward developing MLLMs that balance high-level cognitive depth with manageable computational costs, potentially serving as a reliable framework for real-time multimodal reasoning systems.